\documentclass[letterpaper, 10 pt, conference]{ieeeconf}  

\IEEEoverridecommandlockouts                              

\usepackage{graphics} 
\usepackage{epsfig} 
\usepackage{mathptmx} 
\usepackage{times} 
\usepackage{amsmath} 
\usepackage{amssymb}  
\usepackage{tikz}
\usepackage{algorithm}
\usepackage{algpseudocode}
\usepackage{censor}
\algnewcommand{\LineComment}[1]{\State \(\triangleright\) #1}
\title{\LARGE \bf
Learning to build covering structures with continuous adjustments
}

\author{Gabriel Vallat, Maryam Kamgarpour and Stefana Parascho
}

\begin{document}

\maketitle
\thispagestyle{empty}
\pagestyle{empty}

\begin{abstract}
Robotic construction offers the potential to use materials more efficiently and create complex geometries, but current methods rely on rigid, high-precision plans that cannot accommodate the tolerances, inaccuracies, and unexpected changes inherent in physical fabrication. In this work, we introduce a reinforcement learning approach that forgoes predefined plans entirely, instead generating construction sequences adaptively as the structure is built. Our method operates on graph-structured state representations and a mixed (parameterized) action space, requiring both discrete block selection and continuous placement parameters. Because the stability simulation of a structure is computationally heavy, we develop an efficient exploration strategy by incorporating unilateral edges into graph neural networks, extending soft actor-critic (SAC) to this hybrid setting. We evaluate our algorithm, HSAC, against the prior method hybrid-PPO (HPPO), demonstrating significantly higher asymptotic performance and good sample efficiency. We also demonstrate HSAC's robustness to hyperparameter choices and its exploration capability, handling up to 10 discrete actions without performance degradation. Finally, we validate our approach on a physical two-robot setup, successfully building a spanning arch with 3D-printed blocks in closed-loop execution, confirming that policies trained in simulation transfer to real hardware.
\end{abstract}

\section{Introduction}
\begin{figure}
    \centering
    \includegraphics[width=0.49\linewidth]{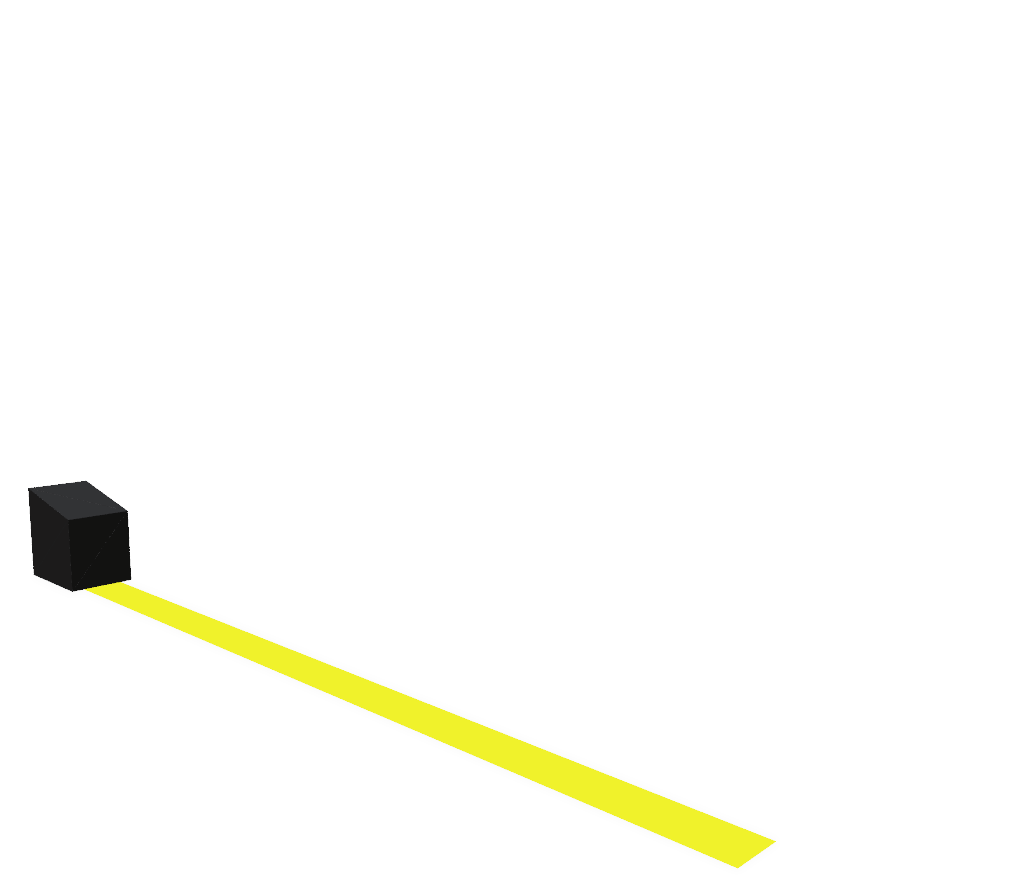}
    \includegraphics[width=0.49\linewidth]{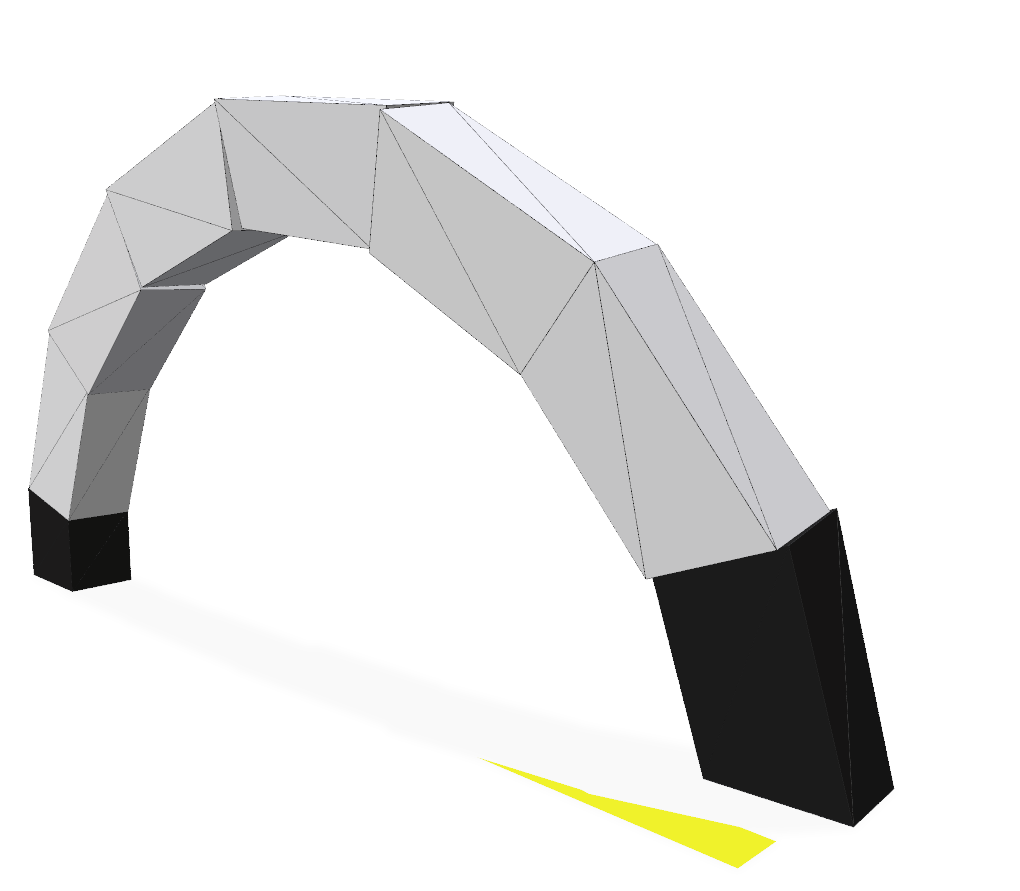}
    \caption{Initial state (left): it contains an initial block and a yellow rectangle that the structure needs to cover. Example of an arch built with small noise added at each step (right): The arch slowly deviates from the desired trajectory, and the yellow area is not completely covered.}
    \label{fig:setup}
\end{figure}

Robotic construction is becoming increasingly prominent, allowing designers and architects to use materials more efficiently and enabling the creation of complex geometries that would be difficult to achieve with traditional methods \cite{Salamanca2023semiramis}. However, this approach still has significant limitations. One specific challenge is the need for materials to conform to a high-precision plan defined before construction begins. Tolerances, inaccuracies, and unexpected changes create mismatches with this idealized case, as illustrated in Fig.~\ref{fig:setup}. In the worst case, such errors can lead to catastrophic failure; at the very least, they necessitate costly and time-consuming plan revisions during construction. To address this, we propose to forego the use of an initial plan entirely and instead generate one incrementally, at each time step of the construction process.

To develop our method, we build upon an assembly of rigid blocks with dry joints. In our experiments, we use plastic blocks, though stone blocks could be used in practice. This fabrication process allows components to be reused but requires the structure to remain purely under compression. After evaluating different construction methods, such as single-robot block stacking, we choose to build an arch using two robots working in tandem \cite{parascho2020vault, Jingwen2023AutonomousConstruction}: one placing blocks while the other provides temporary support. This method requires the negotiation of multiple construction parameters, which is not easily done intuitively, while highlighting the impact of minor placement errors, keeping the structure perpetually on the verge of instability.

To address the challenge above, we model the construction process as a controlled dynamical system and generate a policy: a mapping from states to actions. This formulation ensures fabrication feasibility by considering structural stability at each step of construction. Furthermore, it accommodates continuous adjustments in actions, whereas a plan-based approach typically requires restarting once certain thresholds are exceeded. Despite its promise, research on plan-free methods remains limited; most methods still rely on plans and attempt to increase their adaptability. For instance, some approaches allow the sequence of block placements to change in response to part delivery delays \cite{Ziqi2025Learn2Assemble}.

Previous plan-free strategies also generally rely on strong simplifications, such as discretizing the space \cite{Vallat2023RLspanning,wang2026learningbuildautonomousrobotic} or only considering 2D structures \cite{Bapst2019StructuredConstruction, Jingwen2023AutonomousConstruction}. Given the complexity of our dynamical system, we face the same constraints as these approaches and cannot rely on model-based control frameworks. We therefore turn to model-free reinforcement learning (RL).

In robotic construction, learning a policy requires reasoning over a mixed continuous–discrete action space: the agent must choose both which block to place (a discrete selection from available types) and where to place it (a continuous parameter). These decisions are coupled—the optimal placement depends on the chosen block, and vice versa—and existing algorithms such as HPPO \cite{Fan2019HPPO} and parametrized-DDPG \cite{hausknecht2024DDPGmixed} struggle in complex environments with costly physics-based stability checks. Mixed action spaces of this kind arise broadly in robotics, from manipulation to locomotion, yet remain poorly addressed by current methods.

We address this challenge with hybrid soft actor-critic (HSAC), a novel extension of SAC to mixed action spaces that, like most RL algorithms, makes no assumptions about state representation. For our construction task, we represent the structure as a graph—a problem-specific choice that allows us to leverage geometric invariants (translation, permutation, and unlike prior work \cite{Ziqi2025Learn2Assemble}, vertical rotation) to improve generalization.  This combination of generality and domain-specific inductive bias makes HSAC broadly applicable to robotics problems that couple discrete and continuous decisions.

The resulting algorithm is, to our knowledge, the first to achieve sample-efficient learning in mixed action spaces with costly simulation. We validate HSAC against HPPO \cite{Fan2019HPPO} and demonstrate its use on a physical two-robot setup that builds an arch from 3D-printed blocks, confirming that policies trained in simulation transfer to real hardware.
\section{Dynamical system definition}
As an example of a robotic construction task, we use a spanning structure as shown in Fig.~\ref{fig:setup}. Starting from an initial fixed block, one robot places a new element while another holds the free end to maintain stability. To enable an agent to design a structure autonomously, we require two key elements: a training environment, described in this section as a discrete-time dynamical system, and a learning algorithm, covered in section \ref{sec:algo}, to find an optimal policy for the process.

As the construction of a structure is inherently sequential, we model it as a discrete-time controlled dynamic process, described by $(S,A,T,R,s_0,\gamma)$, where $S$ is the state space, $A$ the action space, $T:S\times A\rightarrow S$ the transition function, $R:S\times A \rightarrow \mathbb{R}$ the reward function, $s_0\in S$ the initial state and $\gamma\in[0,1]$ is the discount factor. We describe below each of its components. Note that while we use a deterministic transition function here, our method readily extends to stochastic dynamics.

\subsection{State}
\begin{figure}
    \centering
    \includegraphics[width=0.49\linewidth]{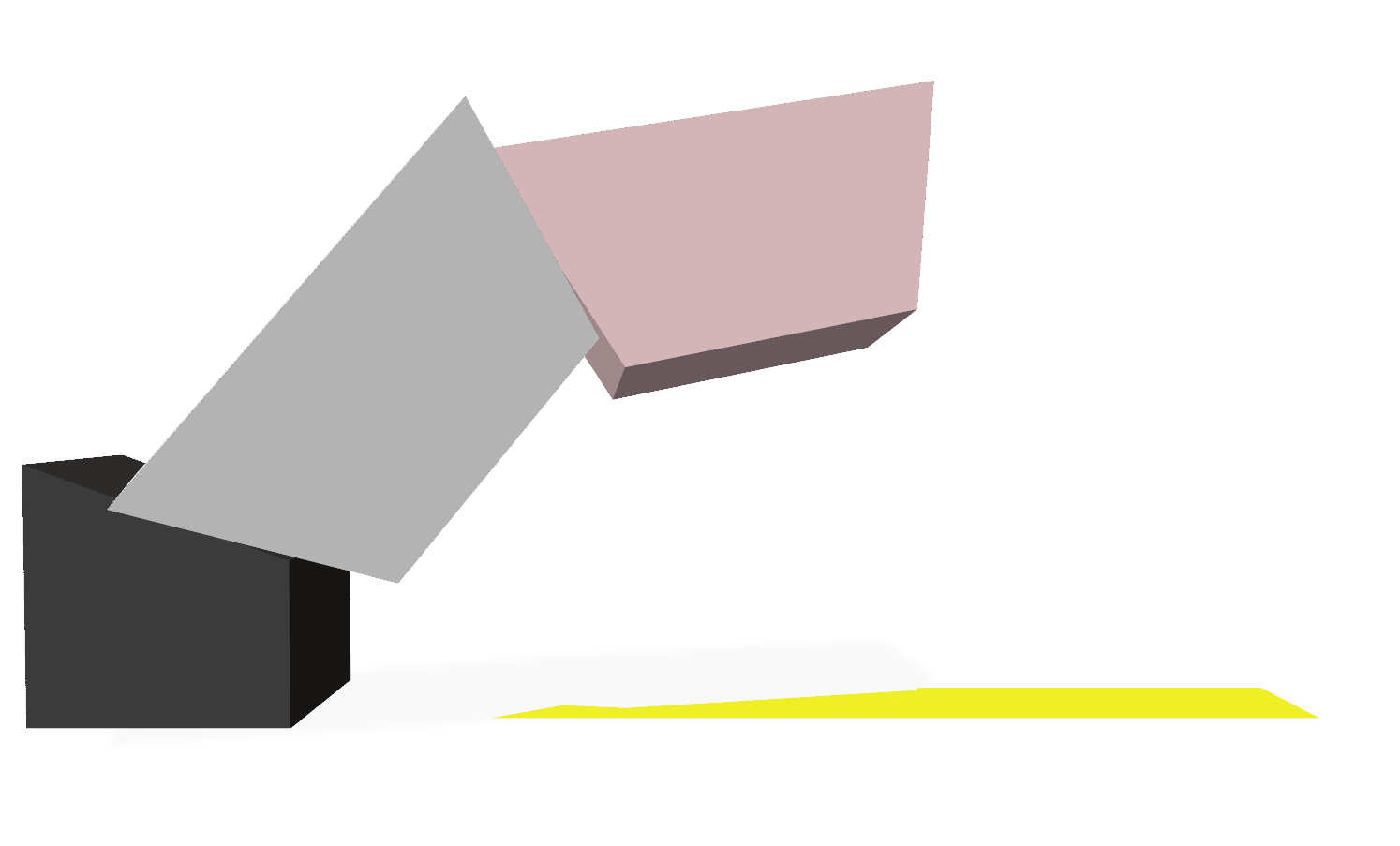}
    \includegraphics[width=0.49\linewidth]{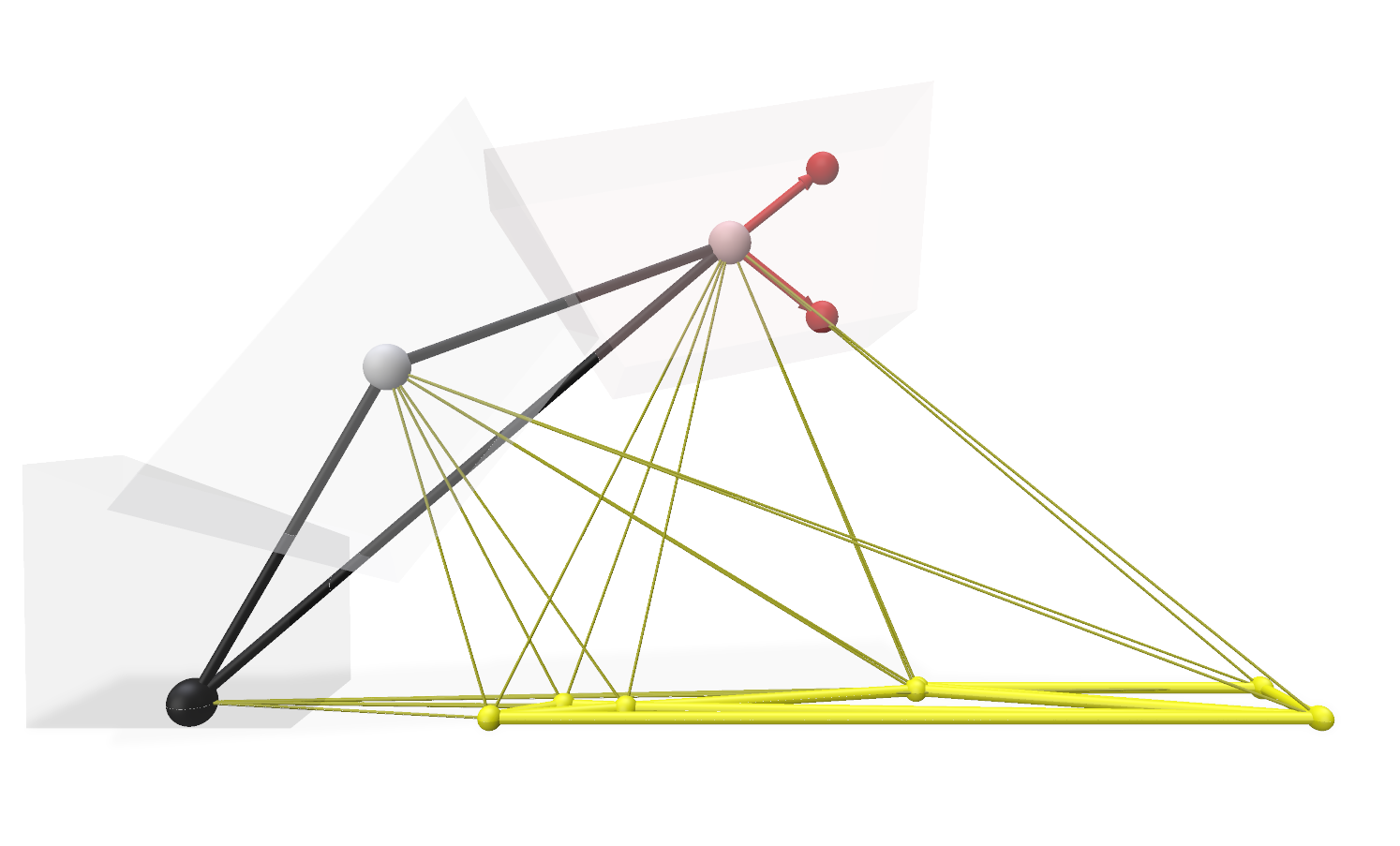}
    \includegraphics[width=0.49\linewidth]{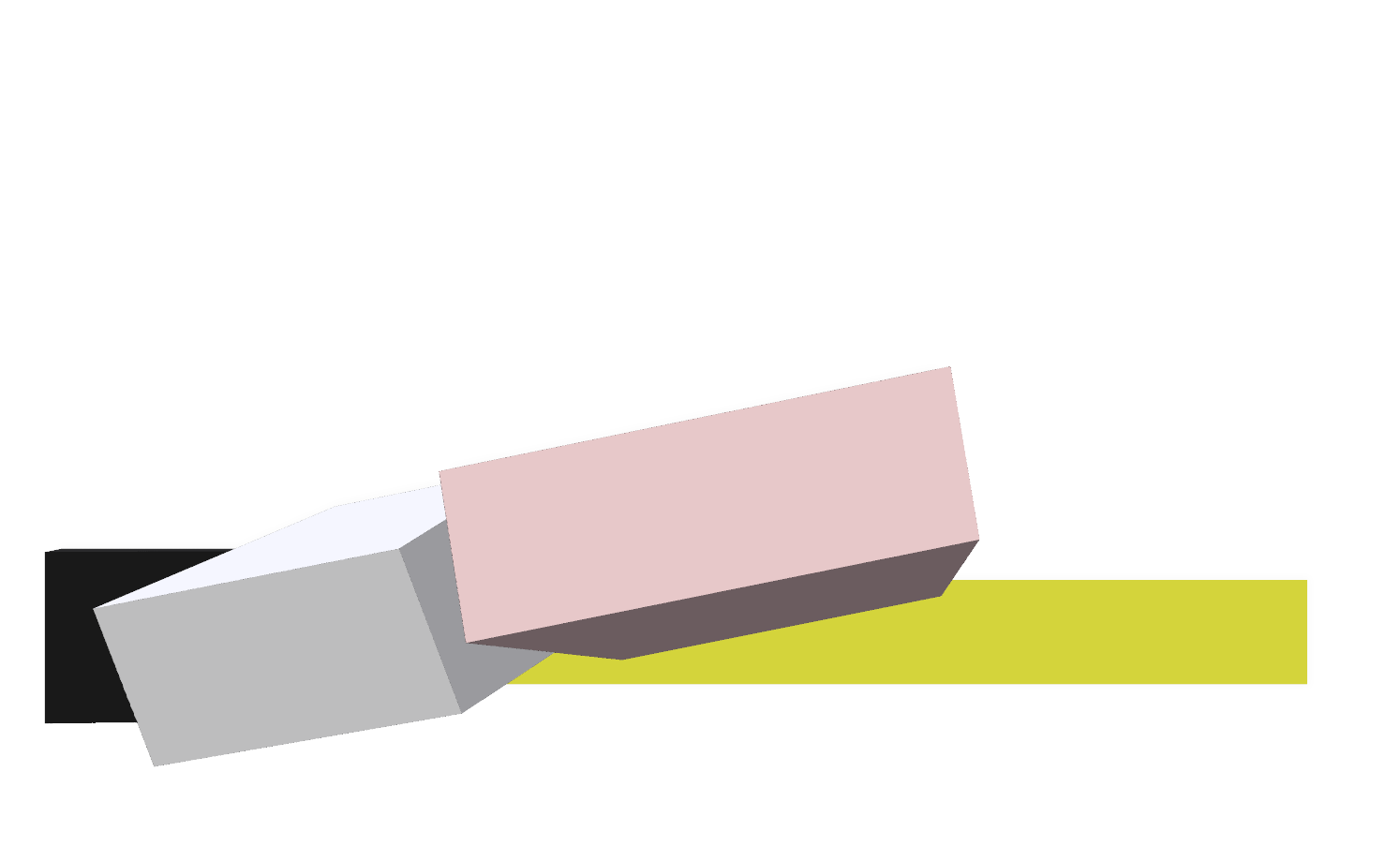}
    \includegraphics[width=0.49\linewidth]{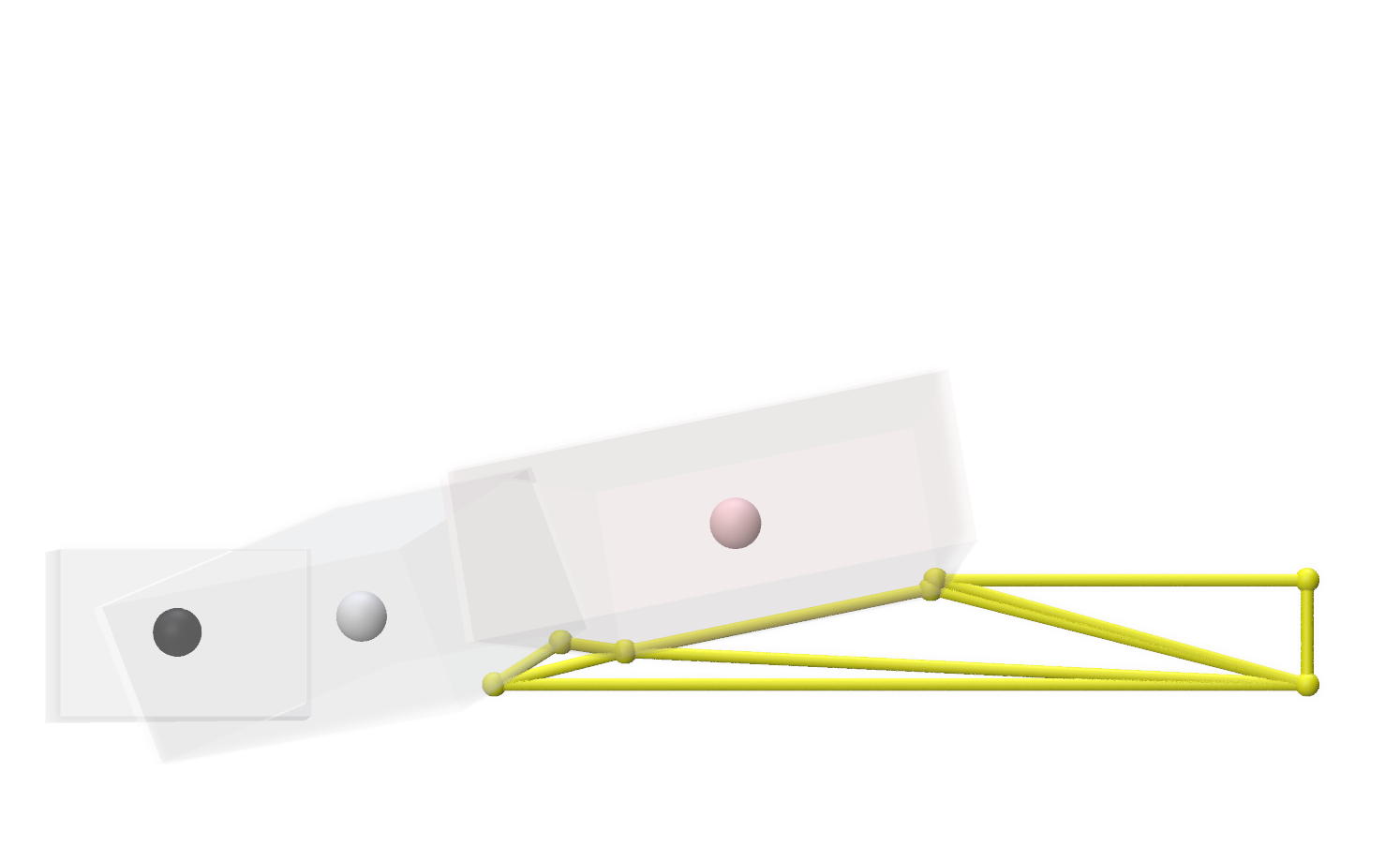}
    \caption{
    Left: a simple three-block structure shown from two perspectives. The black block is the ground, the white block is free-standing, and the red block is held by a robot. Top right: the corresponding graph representation, showing block nodes  $\mathcal{N}_b$, action nodes $\mathcal{N}_a$ (red), and objective $\mathcal{N}_o$ (yellow). Bottom right: detailed view of the objective subgraph, with corner nodes $\mathcal{N}_o$ and the edges $\mathcal{E}_{oo}$ connecting them in yellow.
    }
    \label{fig:graph}
\end{figure}

To train our agent, we need a state $s\in S$ to represent the current structure. Moreover, this state should be invariant to factors that do not modify the optimal action (a change in coordinate system, for example), as our agent would otherwise need to learn to ignore these factors. To fulfill this task, we use a labeled heterogeneous graph $G_s=\{\mathcal{N}_s,\mathcal{E}_s\}$, where $\mathcal{N}_s=\mathcal{N}_{b}\cup\mathcal{N}_{a}\cup\mathcal{N}_{o}$ are the nodes representing, respectively, the present blocks $\mathcal{N}_{b}$, possible type of block to be added $\mathcal{N}_{a}$ and objective $\mathcal{N}_{o}$, and $\mathcal{E}_s = \mathcal{E}_{bb}\cup\mathcal{E}_{ba}\cup\mathcal{E}_{bo}$ are the edges linking these different nodes. The complete process through which the graph is labeled to represent a structure is described in the appendix \ref{app:state}, and an example is shown in Fig.~\ref{fig:graph}. The key novelty of our method is to attach a reference frame to each of our blocks, and to label the edges connecting them with their relative positions, rather than using a global coordinate system. This creates a graph that is invariant to rotation, translation and permutation. Our design follows a simple principle: the state should change only when the optimal action changes. Encoding block positions relative to one another rather than in a global frame achieves this, allowing the agent to learn more efficiently.

\begin{figure}
    \centering
    \includegraphics[width=\linewidth]{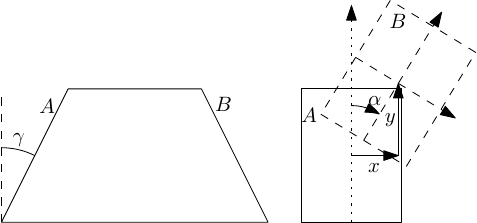}
    \caption{Our blocks are always trapezoidal extrusions, using a fixed aspect ratio, and the angle, or slope, $\gamma$ to parametrize different shapes. On the left, we show how we parametrize the action with the translation $(x,y)$ and the angle $\alpha$. The dashed rectangle is the $B$ face of the next block.}
    \label{fig:trapeze}
\end{figure}

Note also that our state space does not require us to know the exact shape of the blocks, as long as we can simulate its physical behavior: The algorithm is only given a block index that could represent any shape. This framework, with minor adjustments for scanning tolerances, could then be used with any shape, including natural materials with irregular geometries. 

\subsection{Actions}
The action selection proceeds in two steps. First, the agent chooses the type of block $a\in\{1,\dots, N\}$ to add to the structure. Once this decision is made, we describe a target position for the new block. We always orient it so that the face $B$ of the new block touches the face $A$ of the old one. The final position could then be parametrized using the 3 components $u=(x,y,\alpha)$ shown on the right of Fig.~\ref{fig:trapeze}. These represent a 2D translation from the center of face $A$ of the currently supported block to the center of face $B$ of the new block, along with an angle $\alpha$ between the two reference frames. Note that in practice, we use 4 parameters, and not 3, as we represent $\alpha$ by its cosine and sine \cite{Zhou20206Drotation}, creating a continuous representation of the angle.

This makes our continuous action $u$ a vector in $\mathbb{R}^4$, and our complete action set can then be represented as its Cartesian product with the discrete action set, $A=\{1,\dots,N\}\times\cup_{i=1}^N\mathbb{R}^{m_i}$, with $N=3$ and $m_i= m =4$ for all $i$ in our experiments (except if explicitly mentioned). We describe in more detail how we handled this mix of discrete and continuous actions in the section \ref{sec:algo}, dedicated to the algorithm.

\subsection{Initial state}
The initial state $s_0 \in G_s$ simply consists of a permanently fixed base block, and a narrow rectangular area to cover, shown in yellow on Fig.~\ref{fig:setup} and Fig.~\ref{fig:graph}. These two elements allow us to explore different classes of structures. By choosing a small angle $\gamma$ for the initial block, the resulting construction tends to be tall arches, whereas if we increase the length of our rectangular area, we reward the agent for building wider arches.
\subsection{Reward and transition function}
The transition function $T(s,a)$ is separated in two phases. First, the new held block and its contact interface are added to a physics model based on Rigid Block Analysis (RBA) \cite{Ziqi2025Learn2Assemble}, and the previously held block is released. Our model then simulates the structure's stability. If the structure is unstable, the episode ends in failure with a reward of $-1$. Otherwise, the agent receives a reward proportional to the area covered by the new block. This dense reward allows us to train the agent to keep the arch in a chosen plane. Finally, if one of the corners of the newly placed block is below the ground level, we consider the spanning structure as complete and end the episode, marking it as a success.

Together, this dynamical system is particularly hard to optimize: the graph-based state requires specific models (GNNs) to be processed into an action; this action comprises a discrete and a continuous part; and a suboptimal choice can lead to irrecoverable states: either immediate collapse or, worse, guaranteed collapse several steps later. To address these challenges, we propose hybrid soft actor-critic (HSAC), which we describe in detail in the next section.
\section{Hybrid SAC Algorithm}\label{sec:algo}
Classical reinforcement learning algorithms such as proximal policy gradient (PPO) \cite{Schulman2017PPO} and SAC are designed for purely continuous or purely discrete actions and cannot be applied directly. While specialized methods for mixed action spaces exist—HPPO \cite{Fan2019HPPO} being a notable example—they suffer from a critical limitation: their exploration strategy is the same for all states. In contrast, SAC, not yet extended to mixed action sets, uses a state-dependent exploration, empirically leading to both a faster exploration and a higher asymptotic performance. 

In the remainder of this section, we first describe how we adapt the learning algorithm of SAC to accommodate hybrid actions, and then show how we can use GNNs not only to handle the graph-structured states, but also to remove the common over-parametrization of the $Q$-function caused by hybrid actions.
\subsubsection{HSAC}                              
Our training algorithm alternates between two phases. The first one consists of sampling the environment, using a stochastic policy $\pi_\phi((a,u)|s)$, where the discrete action is denoted as $a$ and the continuous one as $u$, and storing the transitions $(s_t,a_t,u_t,r_t,s_{t+1})$ in a large replay buffer $\mathcal{D}$. 
The second phase consists in optimizing the parameters $\phi$ of this policy and three other functions: a $V$-function $V_\psi:S\rightarrow \mathbb{R}$, and two $Q$-functions $Q_{\theta_i}:S\times A \rightarrow \mathbb{R}$, using the double Q-learning method \cite{Fujimoto2018DoubleQ}. We handle the mixed action set by using the chain rule: $\pi_\phi((a,u)|s)=\pi_\phi^c(u|a,s)\pi^d_\phi(a|s)$, and optimize separately the discrete and continuous parts. These functions are optimized using objectives similar to those in the original SAC algorithm, with the loss functions given by:
\begin{align}
    J_Q(\theta_i) &= \mathbb{E}_{s_t,a_t,u_t\sim\mathcal{D}}\left[\frac{1}{2}(Q_{\theta_i}(s_t,a_t,u_t)-\hat{Q}(s_t,a_t,u_t))^2\right],\label{eq:JQ}\\
    J_V(\psi) &= \mathbb{E}_{s_t\sim\mathcal{D},a,u\sim\pi_\phi(s_t)}\biggl[\frac{1}{2}(V_\psi(s_t)-[\min_iQ_{\theta_i}(s_t,a,u)\notag\\
    &-\alpha_c\log \pi_\phi^c(u|a,s_t)-\alpha_d \log(\pi_\phi^d(a|s_t)]))^2\biggr], \label{eq:JV}\\
    J_{\pi^d}(\phi) &=  \mathbb{E}_{s_t\sim \mathcal{D}, u\sim \pi^c_\phi(\cdot|\cdot,s_t)}\biggl[\\ \notag &\text{D}_{KL}\left(\pi^d_\phi(\cdot|s_t)\middle|\middle|\frac{\exp(\alpha_d^{-1}\min_iQ_{\theta_i}(s_t,\cdot,u))}{Z^d_{\theta}(s,u)}\right)\biggr],\label{eq:Jpid}\\
    J_{\pi^c}(\phi)&=\mathbb{E}_{s_t\sim \mathcal{D}, a\sim \pi^d(\cdot|s_t)}\biggl[\\ \notag &\text{D}_{KL}\left( \pi^c_\phi(\cdot|a,s_t)\middle|\middle|\frac{\exp(\alpha_c^{-1}\min_i Q_{\theta_i}(s_t,a,\cdot))}{Z^c_\theta(s,a)}\right)
    \biggr],\label{eq:Jpic}\\
\end{align}
where $\hat Q(s_t,a_t,u_t) = r_t +\gamma V_{\bar{\psi}}(s_{t+1})$, $\text{D}_{KL}$ denotes the KL-divergence, $\bar\psi$ is an exponential moving average of the parameters $\psi$ and $Z^\cdot_\theta$ are normalizing partition functions that do not contribute to the gradient. In our implementation, the coefficients $\alpha_c$ and $\alpha_d$ are dynamic, and the entropies of both policy components converge to target values $H^c_T$ and $H^d_T$ respectively \cite{Haarnoja2019SACtemp}.

To optimize our policy in practice, we use the same reparameterization as in the original SAC paper \cite{Haarnoja2018SAC}. This allows us to write the continuous stochastic policy as a deterministic function $u = f_u(s,a,\epsilon)$, that takes noise $\epsilon\sim N(0,I_m)$, a spherical Gaussian, as input, implicitly defining $\pi_\phi^c(u|a,s)$ as the probability density of $f_u$ under this noise. This allows us to rewrite the total policy cost as 
\begin{align}
    J_\pi(\phi) =& C(J_{\pi^c}(\phi)+ J_{\pi^d}(\phi))& \notag\\=& \mathbb{E}_{s_t\sim \mathcal{D},\epsilon \sim N(0,I_m)}\Biggl[\sum_{a=1}^N \pi^d_\phi(a|s_t)\notag\\ \cdot&\bigl( \alpha_c \log(\pi_\phi^c(f_u(s,a,\epsilon)|a,s_t)) + \alpha_d\log(\pi^d_\phi(a|s_t))\notag \\&- Q(s_t,a,f_u(s,a,\epsilon))\bigr)\Biggr],\label{eq:reparametrized}
\end{align}
where $C$ is constant with respect to $\phi$ and can therefore be ignored and $Q(s,a,u)= \min_iQ_{\theta_i}(s,a,u)$. This yields the update procedure summarized in Algorithm~\ref{alg:HSAC}.
\begin{algorithm}
    \caption{HSAC}\label{alg:HSAC}
\begin{algorithmic}
\Require Some initial parameters $\phi$, $\theta_1$, $\theta_2$ and $\psi=\bar{\psi}$, and an empty replay buffer $\mathcal{D}$.
\State $t\gets 0$
\While{$t< t_{max}$}
\State{$t_{ep}\gets0$}
\State{$s_t\gets s_0$}
\While{Episode not terminated}
    \State{$a_t\sim \pi^d_\phi(\cdot|s_t)$}
    \State{$u_t\sim\pi^c_\phi(\cdot|a_t,s_t)$}
    \State{$\mathcal{D}\gets\mathcal{D}\cup(s_t,a_t,u_t,R(s_t,a_t,u_t),T(s_t,a_t,u_t))$}
    \State{$s_{t+1}\gets T(s_t,a_t,u_t)$}
    \State{$t\gets t+1$}
    \State{$t_{ep}\gets t_{ep}+1$}
\EndWhile
\State{$t_{opt}\gets 0$}
\While{$t_{opt}<k t_{eq}$}
    \State{$\theta_i\gets\theta_i - \delta \hat\nabla_{\theta_i}J_Q(\theta_i)$} \Comment{Using eq.~\ref{eq:JQ}}
    \State{$\psi\gets \psi -\delta \hat\nabla_\psi J_V(\psi)$} \Comment{Using eq.~\ref{eq:JV}}
    \State{$\bar{\psi}\gets \tau \psi + (1-\tau)\bar\psi$}
    \State{$\phi \gets \phi - \delta\nabla J_\pi(\phi)$}\Comment{Using eq.~\ref{eq:reparametrized}}
    \LineComment{Update of $\alpha_c$ and $\alpha_d$}
    \State{Sample $s\sim\mathcal{D}$, $\epsilon_a\sim N(0,I_m)$}
    \State{$H^d\gets\sum_{a=1}^N\pi^d_\phi(a|s)\log(\pi^d_\phi(a|s)))$}
    \State{$H^c \gets\sum_{a}\pi^d(a|s)\log(\pi^c(f(s,a,\epsilon_a)|a,s)))$}
    \State{$\log\alpha_c \gets \log\alpha_c -\delta (H_T^c -H^c)$}
    \State{$\log\alpha_d \gets \log\alpha_d -\delta (H_T^d -H^d)$}
    
    \State{$t_{opt}\gets t_{opt}+1$}
\EndWhile
\EndWhile
\end{algorithmic}
\end{algorithm}

The key novelty of our algorithm is to split the policy cost into Equations~\ref{eq:Jpid} and \ref{eq:Jpic}, instead of directly computing a single cost $J_\pi(\phi)$ for both the continuous and discrete policies. Note that doing so would result in the exact same loss functions as in the original SAC algorithm. Splitting the cost allows us to control the importance of the entropy of each part independently, using different $\alpha_c$ and $\alpha_d$. This method is necessary in practice, as we observed that when using a single coefficient for both parts of the policy, our agents consistently set the discrete entropy close to $0$ (i.e., chose the discrete action nearly deterministically), and used a more stochastic continuous action to reach the target total entropy. This behavior causes the learning to be stuck in local minima, while our method ensures that our discrete policy is explorative enough.
\subsubsection{Graph neural networks}
\begin{figure}
    \centering
    \includegraphics[width=\linewidth]{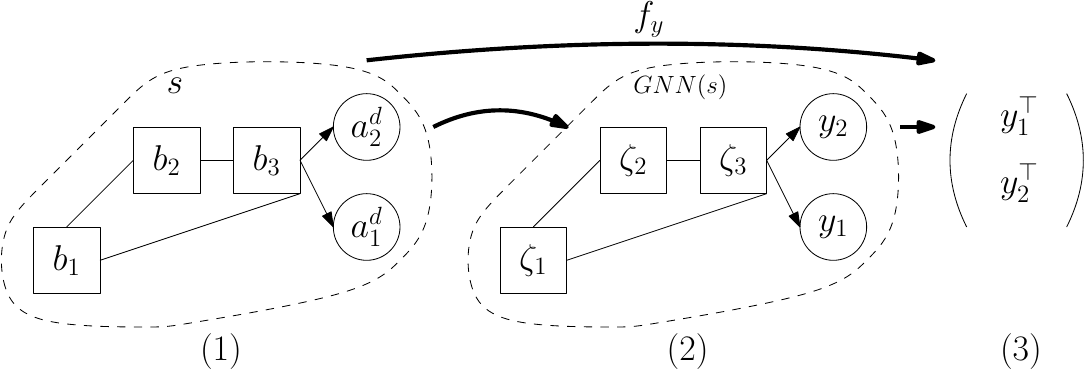}
    \caption{GNN architecture. (1) Input state (the objective nodes are not drawn for simplicity) (2) Output of the GNN. The features attached to the block nodes $\zeta$ can be seen as hidden states of a recurrent neural network and are discarded. (3) Output of the function $f_y(s)$, in this case, a matrix of size $2\times n_y$.}
    \label{fig:GNN}
\end{figure}

Due to the structure of our state and action spaces, we need to parametrize a map from graphs to a mix of continuous and discrete variables. We choose to do so using graph neural networks (GNN) and fully connected networks. The first step of each of these maps is to process heterogeneous graphs to extract features at the discrete action level: $f_{y}: G_s \rightarrow \mathbb{R}^{N\times n_{y}}$, where $n_y$ is the size of the action-level feature. These features are then used to compute the policy and values. We use the transformer convolutional networks \cite{Shi2020TransformerConv} to build our GNN. We then represent $f_y$ by keeping the $N$ features $y_a\in\mathbb{R}^{n_\phi}$, with $ a\in\{1,\dots,N\}$, attached to the action nodes as shown in Fig.~\ref{fig:GNN}. We use this architecture as a basis for both $V_\psi$ and $\pi_\phi$.

To compute $V_\psi$, these features are aggregated using the minimum, maximum, and mean value across the nodes, resulting in a vector $v_\psi\in\mathbb{R}^{3n_\phi}$. This vector is then used as input of a fully connected neural network with only one output. 

To compute the policy, we process each of the features $y_a$ as a separate input of three fully connected neural networks $f_d$, $f_\mu$ and $f_\Sigma$. This allows us to parametrize
\begin{align}
    \pi^d_\phi(a|s)&=\frac{\exp(f_d(y_a))}{\sum_{j=1}^N \exp(f_d(y_j))},\\
    \pi^c_\phi(u|a,s) &= N(f_\mu(y_a),f_\Sigma(y_a)),
\end{align}

where $N(\mu,\Sigma)$ is a multivariate Gaussian distribution of mean $\mu$ and covariance $\Sigma$. Note that the deterministic function $f_u(s,a,\epsilon)$ can then be easily written as $f_u(s,a,\epsilon) = f_\mu(y_a) + f_\Sigma(y_a)\mathbf{\epsilon}$.

Finally, to parametrize an estimation of $Q:S\times A \rightarrow \mathbb{R}$, we take inspiration from deep Q-networks (DQN) \cite{Mnih2015DQN} with a discrete action space of size $N$. In that setting, a neural network $f_{Qd}: S\rightarrow\mathbb{R}^N$ is used to do so, with $Q(s,a)=f_Q(s)_a$ . However, simply extending this method to mixed action sets has led to little success. The authors of \cite{Fan2019HPPO} note that naively expanding DQN to mixed action sets causes over-parametrization problems. As an example, let us define a deep hybrid Q-function $Q_\theta:S\times \mathbb{R}^{Nm}\rightarrow\mathbb{R}^N$, with $N=2$. We can then write
\begin{align}
    Q_\theta(s,\begin{bmatrix} u_{1} \\u_{2}\end{bmatrix}) &= \begin{bmatrix}q_1 \\q_2
    \end{bmatrix},\\
    Q_\theta(s,\begin{bmatrix} u_{1} \\u_{2}'\end{bmatrix}) &= \begin{bmatrix}q_1' \\q_2'
    \end{bmatrix}\label{eq:Qindep},
\end{align}
where $u_2$ and $u_2'$ are two different actions parameters for discrete action $a=2$. If we parametrize $Q_\theta$ using a fully connected network, we cannot ensure that $q_1 = q_1'$.

\begin{figure}
    \centering
    \includegraphics[width=0.75\linewidth]{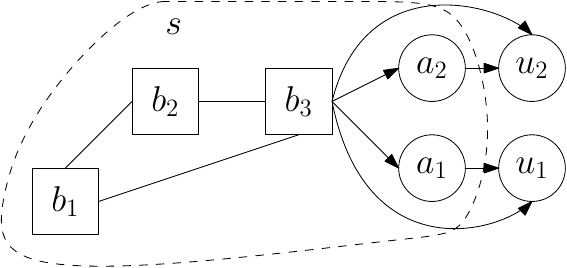}
    \caption{The actions are attached to the state with unidirectional edges, ensuring that $q_1=q_1'$ in Equation \ref{eq:Qindep}. In this example, the state is composed of 3 blocks, and the two possible block types that can be added. The objective nodes are omitted for simplicity. }
    \label{fig:GNNQ}
\end{figure}

The key novelty of our method—and what we believe is the main reason for its effectiveness—is leveraging unidirectional edges in the input graph, as shown in Fig.~\ref{fig:GNNQ}, to enforce this equality. As $u_1$ is not connected to $u_2$, we structurally ensure that the two variables do not interact. While we use this architecture to adapt SAC to our setup, we believe that this method can be used to extend any DQN-based method to mixed action sets effectively.

\section{Implementation}
\subsection{Simulation}
The code for our environment implementation and algorithm is available on GitHub\footnote{https://github.com/sycamore-lab-EPFL/HSAC}. We used Warp \cite{warp2022} to parallelize contact detection across multiple simulation threads, significantly accelerating training. Gurobi \cite{gurobi} served as our optimization solver for the Rigid Block Analysis (RBA) stability checks. We stored each state-action pair using PyTorch Geometric objects \cite{fey2025pyg}, which enabled efficient batching and GPU-based graph neural network training.
\subsection{Simulation Results}
To test the limits of our algorithm, we performed three experiments. First, we compared the learning curve of our algorithm with HPPO. Then, we showed that our method works with a wide range of hyperparameters. Finally, we illustrated the exploration capabilities and generality of our method by training it in different environments, varying both the friction coefficient between the blocks and the number of discrete actions.
\subsubsection{HSAC vs HPPO}
To benchmark our algorithm, we implemented HPPO using a similar GNN architecture. The hyperparameters were tuned manually, and their values are available in appendix \ref{app:hyperparameters}. As shown in Fig.~\ref{fig:exp1}, HSAC achieves better policies when converged. What the graph does not show, however, is that HSAC requires more time to optimize the network: a single HSAC step takes on average five times longer than an HPPO step. This overhead can be mitigated by adjusting the ratio of training updates to simulation steps, effectively making this ratio a tunable hyperparameter that can be optimized based on the available simulation budget.
\begin{figure}
    \centering
    \includegraphics[width=\linewidth]{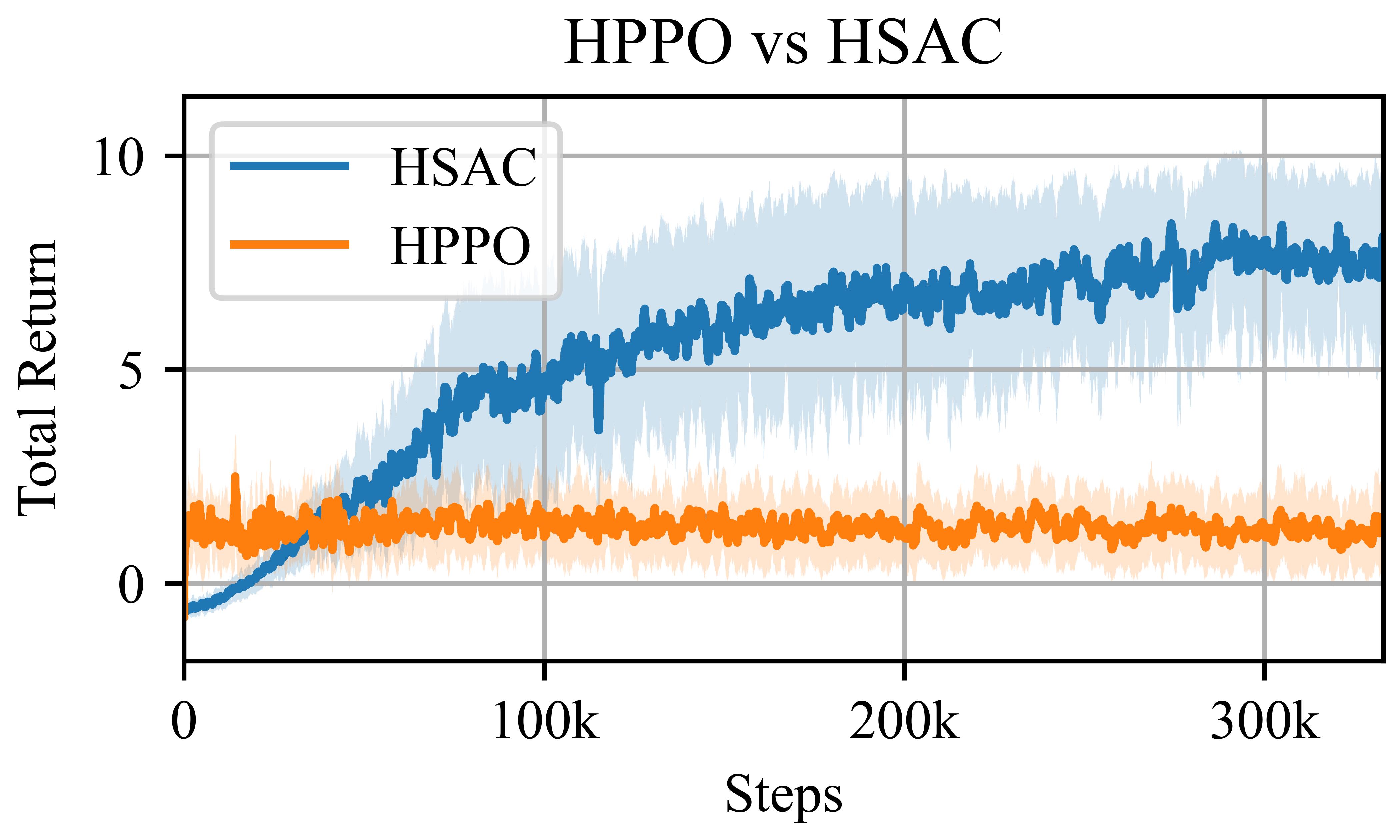}
    \caption{Total reward accumulated during an episode. Results are averaged over three runs. As one can see, HPPO gets stuck in local minima and never reaches the maximum return.}
    \label{fig:exp1}
\end{figure}
\subsubsection{Hyperparameters}
\begin{figure}
    \centering
    \includegraphics[width=\linewidth]{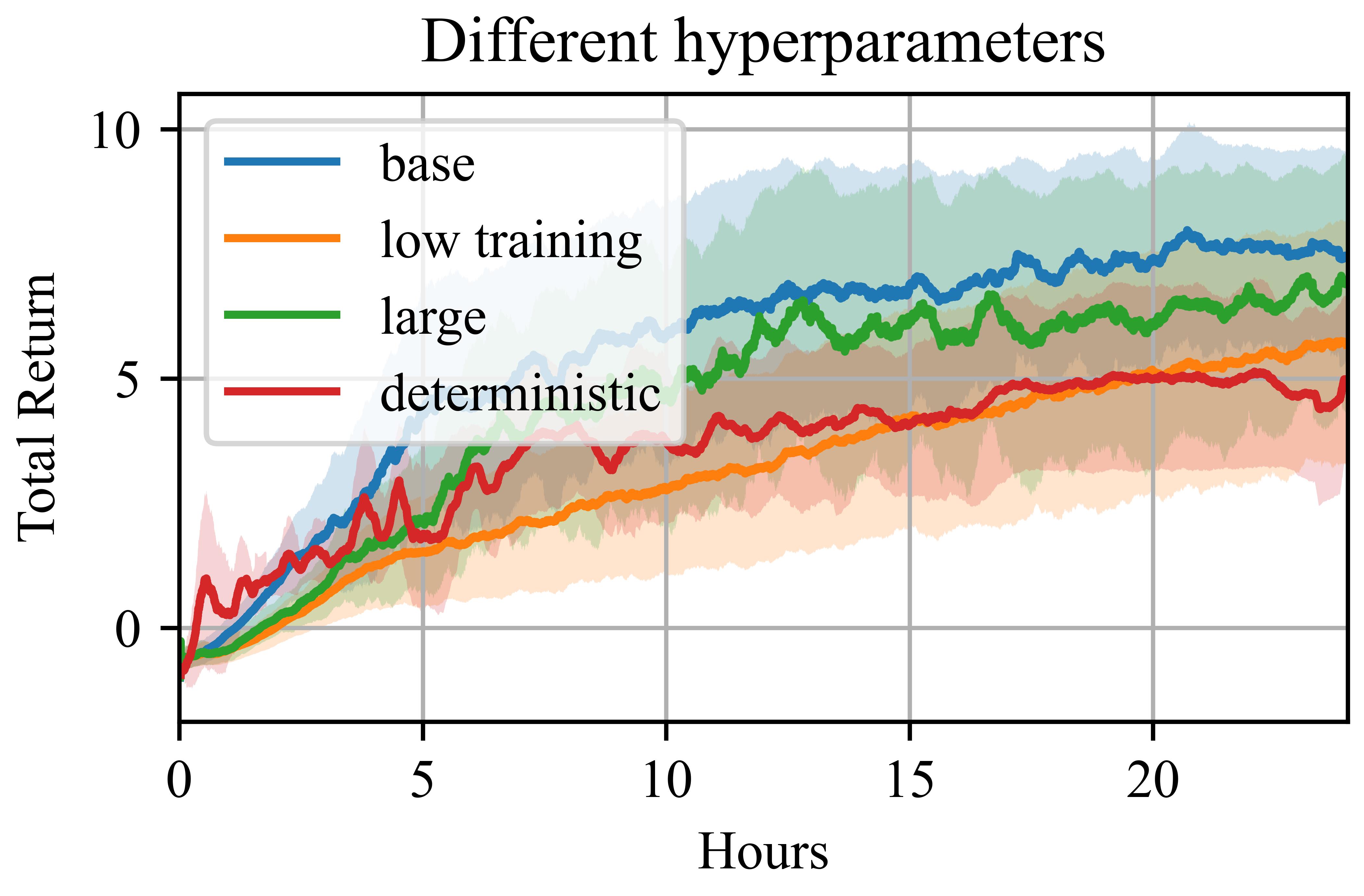}
    \caption{Total reward accumulated during an episode. We used time rather than samples, as the duration of a single step could vary widely. Results are averaged over three runs.}
    \label{fig:exp2}
\end{figure}

To show the stability of HSAC with respect to hyperparameters, we ran four different versions.
\begin{itemize}
    \item A base setup, using around 24 million parameters and performing $64$ update steps for each environment step taken
    \item A larger model, using around 143 million parameters, still performing $64$ update steps per environment step
    \item A less trained model, using  24 million  parameters and performing $8$ update steps per environment step
    \item A deterministic rollout, using the same training setup as the base one, but using a deterministic policy when interacting with the environment
\end{itemize}
As shown in Fig.~\ref{fig:exp2}, using a larger model did not significantly improve the results, and a smaller training ratio slowed convergence. Moreover, as the deterministic version of the algorithm cannot take advantage of the parallelization of the environments, its performance is slightly worse than that of the stochastic policy.
\subsubsection{Change in environment}
\begin{figure}
    \centering
    \includegraphics[width=\linewidth]{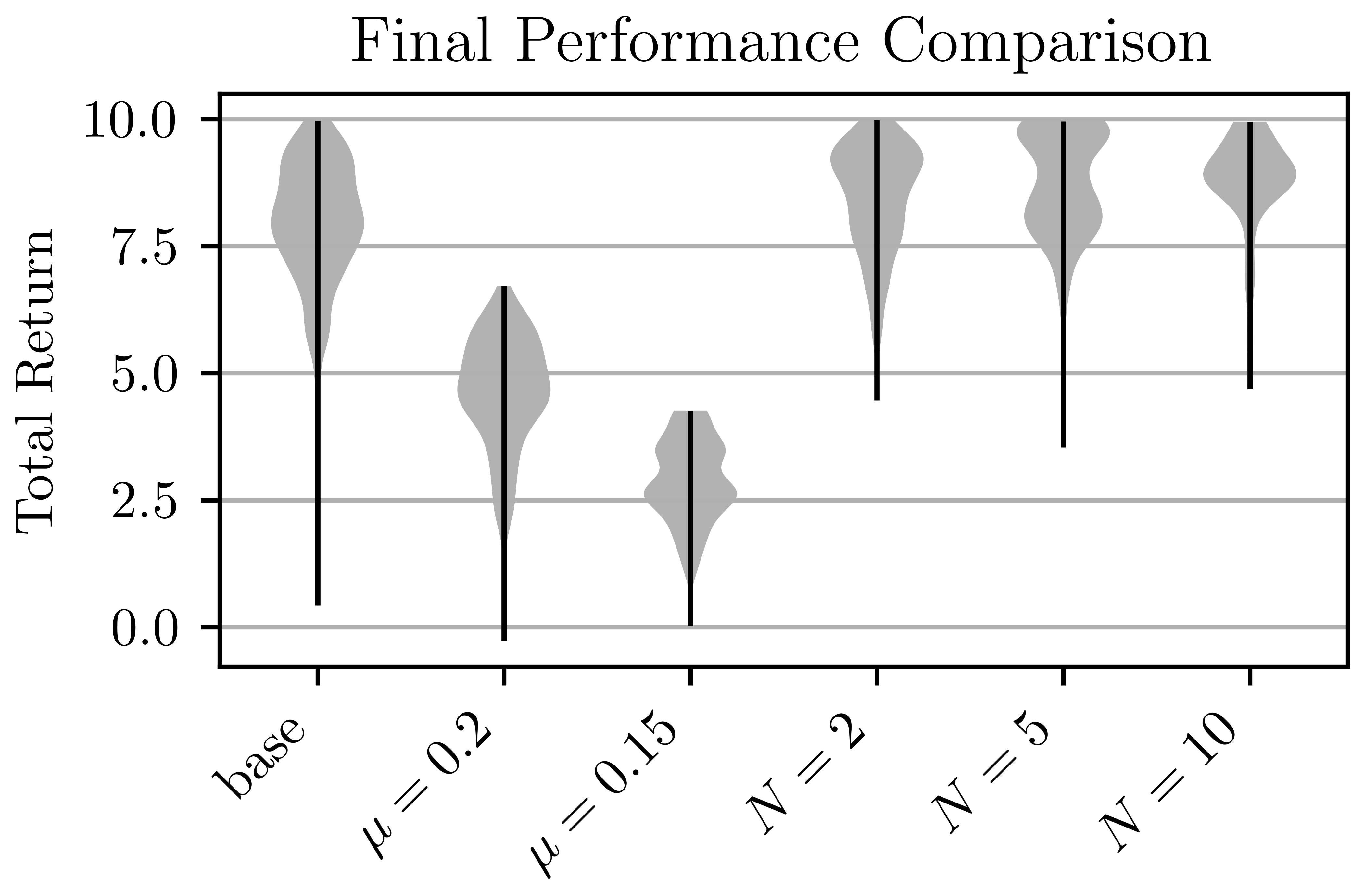}
    \caption{Violin plot of the return over the last 1024 episodes of training, aggregated across 3 runs.}
    \label{fig:exp3}
\end{figure}

To show that our algorithm is able to handle different kinds of environments, we changed some parameters in the simulations. We focused on two parameters: the number of discrete actions, and the friction coefficient between our blocks. An increase in the number of discrete actions $N$, allowing our agent to select more types of blocks, increases the size both of the search space and of the solution space. Increasing the friction coefficient has roughly the same effect, as more states are stable. This setup increases the variety of possible spanning structures, but also allows the agent to search in a bad direction for longer. As shown in Fig.~\ref{fig:exp3}, HSAC can still handle a discrete action set of size $N=10$ effectively, and is able to obtain better results when the friction coefficient is higher. This latter result implies that our algorithm is exploring the environment efficiently, and that narrowing the search space is not required.
\subsubsection{Advantage of a policy}
\begin{figure}
    \centering
    \includegraphics[width=0.75\linewidth]{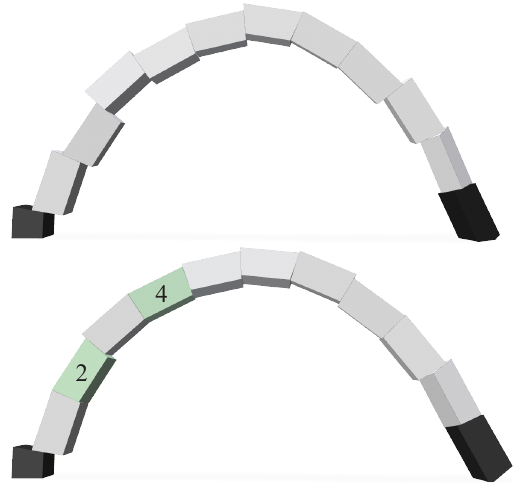}
    \caption{The top arch is the result of our policy. For the lower arch, we forced a different action on steps 2 and 4. Such adaptations are often needed in robotic construction, as the simulation generally does not include all the constraints like temporary obstacles or human intervention.}
    \label{fig:policyadvantage}
\end{figure}

To illustrate how our policy can adapt to unforeseen modifications, we forced the actions taken at the second and fourth steps during construction of the arch shown in Fig.~\ref{fig:policyadvantage}. The top arch is purely produced by a trained policy, and the lower one is the resulting structure after these modifications. As one can see, the agent has been able to compensate for the change over time, leading to nearly identical structures. 
\subsection{Prototype}
\begin{figure*}
    \centering
    \includegraphics[width=\linewidth]{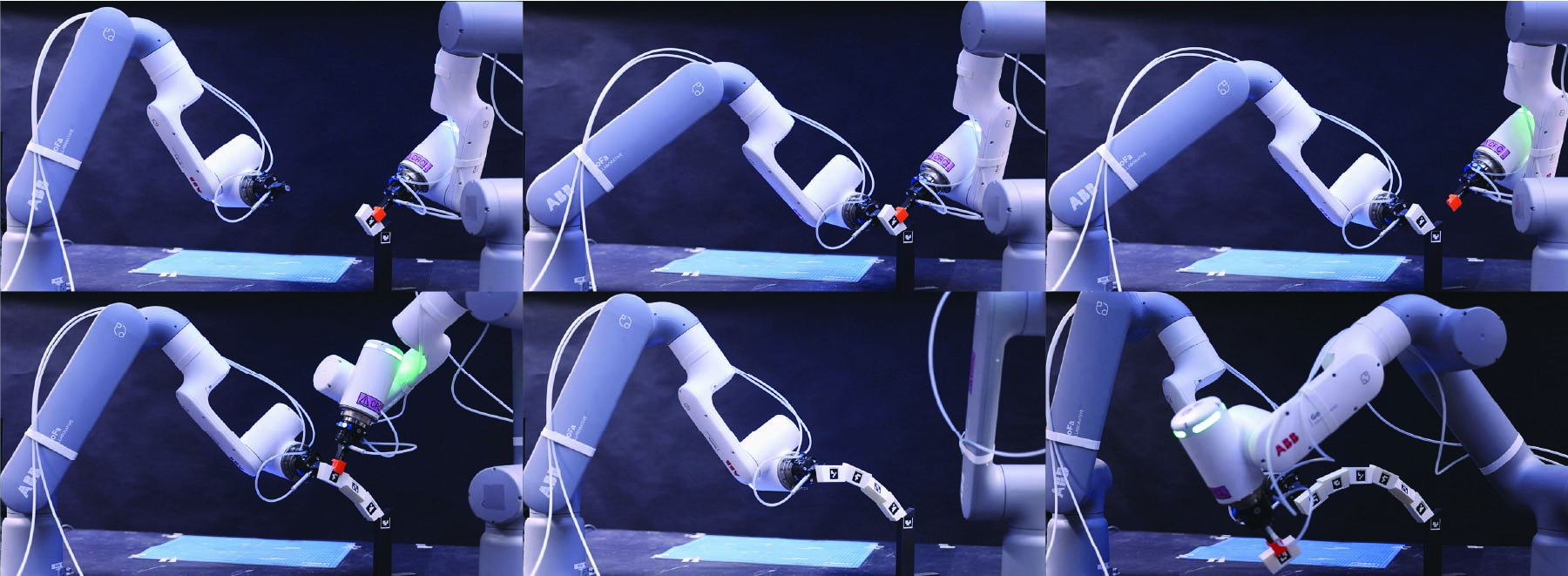}
    \caption{Construction process using two robots. First row (left to right): the first robot places a block, the second robot supports it, and the first robot withdraws. After each placement, the camera captures the block positions, and our policy computes the next block's target location. Second row: subsequent stages of the construction process.}
    \label{fig:construction}
\end{figure*}

As a proof of concept for our algorithm, we used a policy trained with a friction coefficient of $0.15$ to build an arch. We used trapezoid blocks with angles $\gamma=5^\circ,10^\circ$, and $20^\circ$, and a ground block with $\gamma=30^\circ$. Interestingly, our agent did not use any of the $20^\circ$ blocks in its structure.  Our setup, building on the one used in \cite{wang2026learningbuildautonomousrobotic}, consisted of two ABB Gofa robots, a Zivid 3D camera and a set of 3D-printed blocks. At each time-step, the position of each block was captured using the camera and Aruco codes. The position of each block relative to the ground was then transferred to our simulator, and we selected the action based on our trained policy. 
To minimize the influence of calibration errors, we placed all blocks using the same robot, using the other only as a temporary support, leading to the construction process shown in Fig.~\ref{fig:construction}.

\begin{figure}
    \centering
    \includegraphics[width=0.4\linewidth]{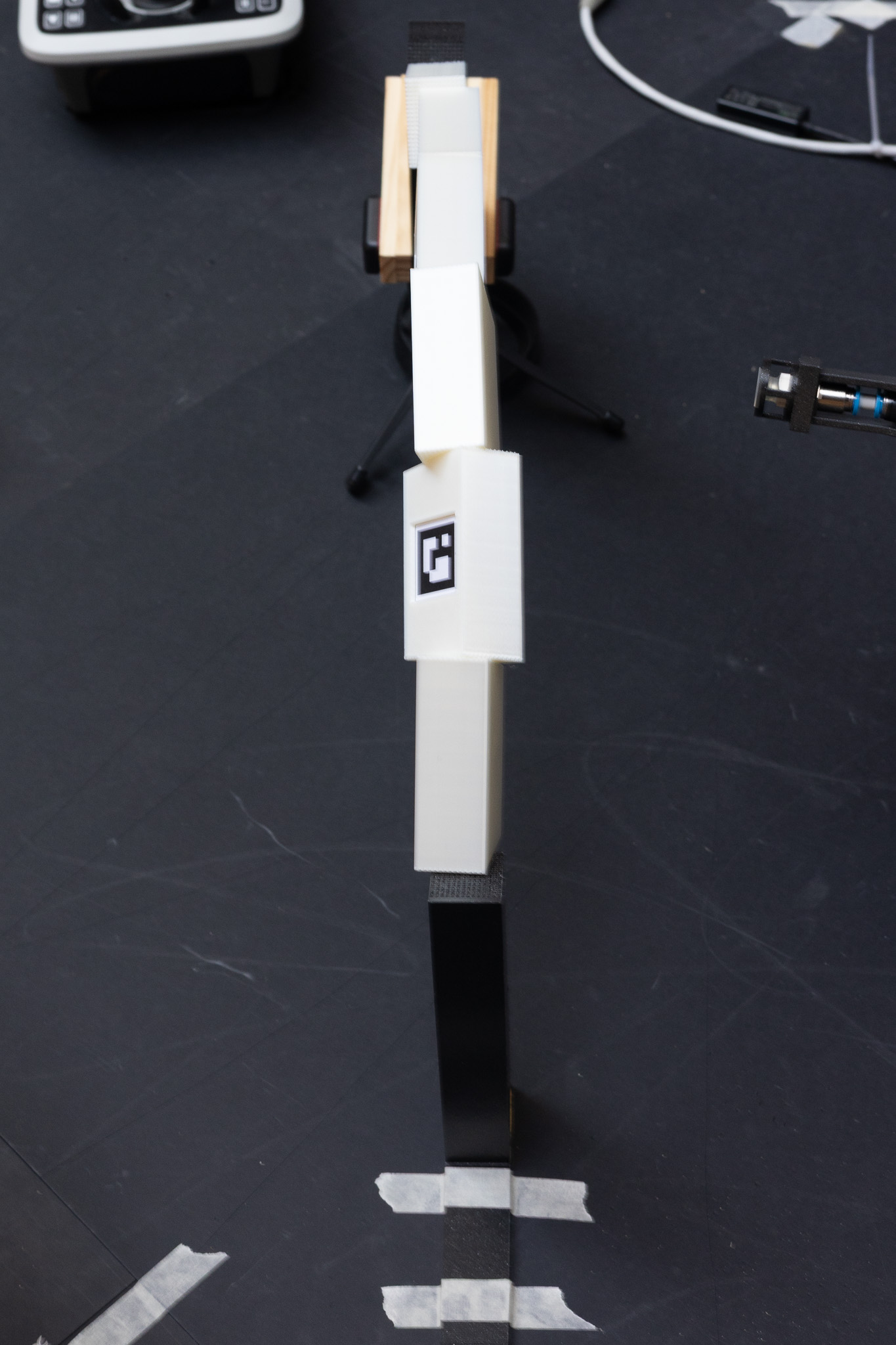}
    \includegraphics[width=0.4\linewidth]{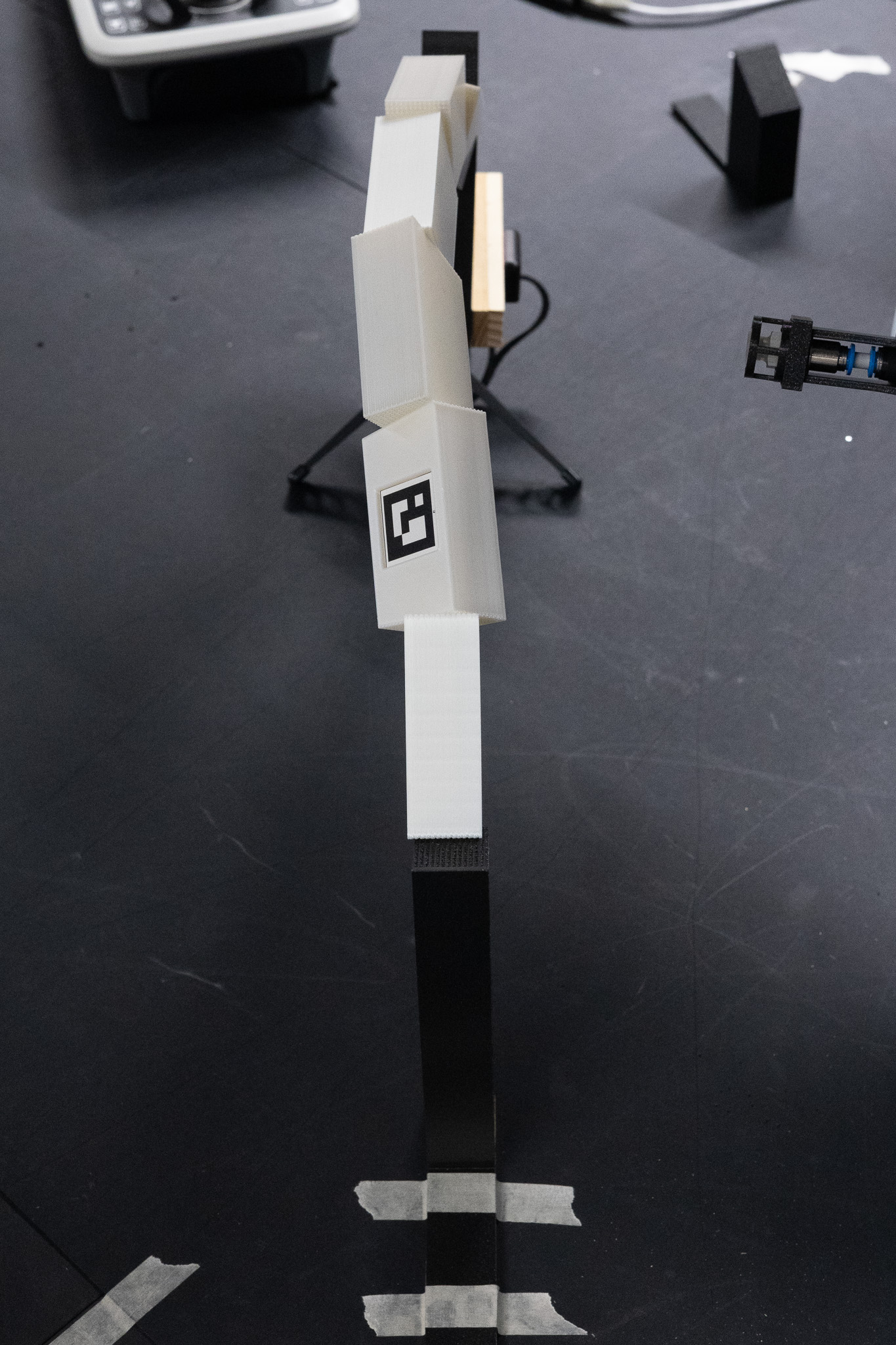}
    \caption{Left: Structure built in closed-loop. Right: Structure built in open-loop. As one can see, the closed-loop is better aligned with the ground block.}
    \label{fig:openvsclosed}
\end{figure}

Finally, we compare the closed-loop system results with those of an open-loop implementation. In this baseline setup, we computed the whole structure in simulation, and used the result as a plan. Our results show that our method was able to keep the arch in the target plane, as shown in the left of Fig.~\ref{fig:openvsclosed}, while the open-loop version deviated more, as shown on the right photo.

\section{Conclusion}
In this work, we introduced a reinforcement learning framework for plan-free robotic construction that operates directly on graph-structured representations of 3D structures. By modeling the construction process as a dynamical system, we formulated the problem as learning a policy that maps states to mixed continuous–discrete actions, selecting both the type of block to place and its precise position. 

The key technical novelty of our approach lies in leveraging graph neural networks with unidirectional edges to enforce independence between Q-values for different discrete actions, a structural inductive bias that proved essential for stable learning. This led to hybrid soft actor-critic (HSAC), an algorithm that extends SAC to mixed action spaces while maintaining adaptive exploration through separate entropy coefficients for the discrete and continuous components of the policy.

Through extensive experiments, we demonstrate that HSAC consistently outperforms prior methods such as HPPO, achieving higher asymptotic performance. We also show its robustness to hyperparameter choices, and that it scales effectively to larger action spaces—handling up to ten discrete block types without degradation—and adapts to varied environment dynamics such as changes in friction coefficients. As a proof of concept, we validated our approach on a physical robot setup, where two ABB GoFa arms successfully built an arch using 3D-printed blocks in closed-loop execution, demonstrating that policies trained in simulation transfer to real hardware.
\addtolength{\textheight}{-0cm}   



\section*{APPENDIX}
\subsection{State Construction} \label{app:state}
\begin{itemize}
    \item The nodes in $\mathcal{N}_b$ are labeled with all features necessary to describe a block, except their position. This includes the type of block used, its $I_{zz}$ inertia component (which encode its vertical orientation), and two binary flags indicating whether the block is held by a robot or part of the fixed setup. Formally, we define a labeling function $f_b:\mathcal{N}_b\rightarrow \{1,\dots,N\}\times \mathbb{R}\times \{0,1\}^2$, where $N$ denotes the number of different block types.
    \item The nodes in $\mathcal{N}_a=\{1,\dots,N\}$ contain the potential blocks that can be added to the structure. While it is not classical to include the list of all possible actions in the state of a dynamical system, these nodes serve as placeholders for the output of the policy model. They are simply labeled with their index, allowing our policy to later differentiate between them. In different setups, where some blocks are only available in finite quantity, we could also remove some nodes from $\mathcal{N}_a$ to restrict the agent to only use available parts.
    \item The nodes in $\mathcal{N}_o$ are used to define the area that the agent has to cover. Each of them represents the corner of a polygon. They do not require a label, as the edges $\mathcal{E}_{oo}$ are sufficient to define any polygon. 
    \item  The edges in $\mathcal{E}_{bb}=\{(i,j):i,j\in\mathcal{N}_b\}$ are connecting all nodes in $\mathcal{N}_b$. Their labels are used to encode the transformation between the reference frames of blocks $i$ and $j$. The resulting labeling function can be defined as $f_{bb}:\mathcal{N}_b\times\mathcal{N}_b\rightarrow\mathbb{R}^{12}$, as we vectorize only the non-trivial components of the $4\times4$ transformation matrix. Note that we keep all $9$ components of the rotation matrix, resulting in an overparameterized representation. The reason behind this choice is to increase the relative number of parameters related to the block position, adding a small inductive bias toward these features in our models.
    \item The edges in $\mathcal{E}_{ba}=\{(i,j):i =\max\mathcal{N}_b,j\in\mathcal{N}_a\}$ attach the most recent block to each action. This instructs the model that the following block will be placed against it. As our models do not modify the labels attached to the edges, we also label these edges with the action they are attached to: $f_{ba}(i,j)=f_a(j)=j$. In practice, this acts similarly to a residual connection in a convolutional neural network, reducing the gradient vanishing problem.
    \item To determine the existence of an edge in $\mathcal{E}_{oo}\subseteq\{(i,j):i,j \in \mathcal{N}_o\}$, we performe a constrained Delaunay triangulation of the polygon we want to cover. The corners of each triangle are then connected. These edges are then labeled by the distance between the two corners. 
    \item Finally, each objective node is linked to each block by the bidirectional edges $\mathcal{E}_{bo}$. Each of the edges is labeled with the position of the corner in the reference frame of the block. 
\end{itemize}
\subsection{Hyperparameters}\label{app:hyperparameters}
\begin{table}[]
    \centering
    \begin{tabular}{c|c}
        Parallel environments & 16 \\
        Discount factor & 0.9 \\
        angles & $5^\circ$, $10^\circ$ \& $20^\circ$ \\
        ground angle & $5^\circ$\\
        friction coef. & 0.3 \\
        
    \end{tabular}
    \caption{Environment base setup (all experiments)}
    \label{tab:env-params}
\end{table}
\begin{table}[]
    \centering
    \begin{tabular}{c|c}
        hidden channels & 64 \\
        GNN layers & 8 \\
        Fully connected layers & 2 \\
        Attention heads & 2 \\
        optimizer & Adam \\
        learning rate & $5\cdot 10^{-4}$\\
        Exponential rate of $\bar{\psi}$ & $10^{-5}$\\
        Replay buffer size & 100'000\\
        Update steps / env steps & 4 \\
        Update batch size & 256 \\
        Continuous target entropy & -3\\
        Discrete target entropy & 0.5 \\
        \end{tabular}
    \caption{HSAC base setup (all experiments)}
    \label{tab:HSAC-params}
\end{table}
\begin{table}[]
    \centering
    \begin{tabular}{c|c}
        $\epsilon$ & 0.2 \\
        GAE $\lambda$ & 0.95 \\
        Continuous entropy bonus & $10^{-3}$\\
        Discrete entropy bonus & $10^{-2}$\\
        
    \end{tabular}
    \caption{HPPO specific parameters}
    \label{tab:HPPO-modifs}
\end{table}
Table~\ref{tab:env-params} lists the parameters of our base training environment, Table~\ref{tab:HSAC-params} those of our base HSAC algorithm, and Table~\ref{tab:HPPO-modifs} the additional hyperparameters for HPPO. For the larger model, we increased the GNN layers to 12 and hidden channels to 128. To reduce training time, we decreased the batch size to 128 and performed only two update steps. For experiments with fewer discrete actions, we retained only blocks with $\gamma = 5^{\circ}$ and $10^{\circ}$; for more actions, we used homogeneous spacing between $\gamma = 5^{\circ}$ and $20^{\circ}$.
\section*{ACKNOWLEDGMENT}
We would like to deeply thank Jingwen Wang for her code to control the robots in close-loop, in particular for her modules regarding path and motion planning and force control.
This work was supported as a part of NCCR Automation, a National Centre of Competence in Research, funded by the Swiss National Science Foundation (grant number 51NF40\_225155), by the EPFL AI center, and by the SNFS (grant number:
\bibliographystyle{IEEEtran}
\bibliography{biblio.bib}

@misc{Schulman2017PPO,
      title={Proximal Policy Optimization Algorithms}, 
      author={John Schulman and Filip Wolski and Prafulla Dhariwal and Alec Radford and Oleg Klimov},
      year={2017},
      eprint={1707.06347},
      archivePrefix={arXiv},
      primaryClass={cs.LG},
      url={https://arxiv.org/abs/1707.06347}, 
}

@inproceedings{Bapst2019StructuredConstruction,
  title={Structured agents for physical construction},
  author={Victor Bapst and Alvaro Sanchez-Gonzalez and Carl Doersch and Kimberly L. Stachenfeld and Pushmeet Kohli and Peter W. Battaglia and Jessica B. Hamrick},
  booktitle={International Conference on Machine Learning},
  year={2019},
  url={https://api.semanticscholar.org/CorpusID:102352078}
}

@INPROCEEDINGS {Zhou20206Drotation,
author = { Zhou, Yi and Barnes, Connelly and Lu, Jingwan and Yang, Jimei and Li, Hao },
booktitle = { 2019 IEEE/CVF Conference on Computer Vision and Pattern Recognition (CVPR) },
title = {{ On the Continuity of Rotation Representations in Neural Networks }},
year = {2019},
volume = {},
ISSN = {},
pages = {5738-5746},
doi = {10.1109/CVPR.2019.00589},
url = {https://doi.ieeecomputersociety.org/10.1109/CVPR.2019.00589},
publisher = {IEEE Computer Society},
address = {Los Alamitos, CA, USA},
month =Jun}

@misc{Ziqi2025Learn2Assemble, title={Learning to Assemble with Alternative Plans}, volume={44}, ISSN={0730-0301}, url={https://infoscience.epfl.ch/handle/20.500.14299/252688}, DOI={10.1145/3730824}, number={4}, publisher={Association for Computing Machinery (ACM)}, author={Wang, Ziqi and Liu, Wenjun and Wang, Jingwen and Vallat, Gabriel and Shi, Fan and Parascho, Stefana and Kamgarpour, Maryam}, year={2025}, month={jul}, pages={1–16}, language={en} }

@inproceedings{Haarnoja2018SAC,
author = {Tuomas Haarnoja and Aurick Zhou and Pieter Abbeel and Sergey Levine}, editor = {Jennifer G. Dy and Andreas Krause},  title = {Soft Actor-Critic: Off-Policy Maximum Entropy Deep Reinforcement Learning with a Stochastic Actor},  booktitle = {Proceedings of the 35th International Conference on Machine Learning,  {ICML} 2018, Stockholmsm{\"{a}}ssan, Stockholm, Sweden, July 10-15, 2018},  series = {Proceedings of Machine Learning Research},  volume = {80},  pages = {1856--1865},  publisher = {PMLR},  year = {2018},  url = {http://proceedings.mlr.press/v80/haarnoja18b.html}}

@misc{Haarnoja2019SACtemp,
      title={Soft Actor-Critic Algorithms and Applications}, 
      author={Tuomas Haarnoja and Aurick Zhou and Kristian Hartikainen and George Tucker and Sehoon Ha and Jie Tan and Vikash Kumar and Henry Zhu and Abhishek Gupta and Pieter Abbeel and Sergey Levine},
      year={2019},
      eprint={1812.05905},
      archivePrefix={arXiv},
      primaryClass={cs.LG},
      url={https://arxiv.org/abs/1812.05905}, 
}

@inproceedings{Fan2019HPPO,
author = {Fan, Zhou and Su, Rui and Zhang, Weinan and Yu, Yong},
title = {Hybrid actor-critic reinforcement learning in parameterized action space},
year = {2019},
isbn = {9780999241141},
publisher = {AAAI Press},
booktitle = {Proceedings of the 28th International Joint Conference on Artificial Intelligence},
pages = {2279–2285},
numpages = {7},
location = {Macao, China},
series = {IJCAI'19}
}

@article{parascho2020vault,
author = {Parascho, Stefana and Han, Isla and Walker, Samantha and Beghini, Alessandro and Bruun, Edvard and Adriaenssens, Sigrid},
year = {2020},
month = {12},
pages = {},
title = {Robotic vault: a cooperative robotic assembly method for brick vault construction},
volume = {4},
journal = {Construction Robotics},
doi = {10.1007/s41693-020-00041-w}
}

@inproceedings{Vallat2023RLspanning,
author = {Vallat, Gabriel and Wang, Jingwen and Maddux, Anna and Kamgarpour, Maryam and Parascho, Stefana},
title = {Reinforcement learning for scaffold-free construction of spanning structures},
year = {2023},
isbn = {9798400703195},
publisher = {Association for Computing Machinery},
address = {New York, NY, USA},
url = {https://doi.org/10.1145/3623263.3623359},
doi = {10.1145/3623263.3623359},
booktitle = {Proceedings of the 8th ACM Symposium on Computational Fabrication},
articleno = {12},
numpages = {12},
location = {New York City, NY, USA},
series = {SCF '23}
}

@inproceedings{Fujimoto2018DoubleQ,
  title={Addressing Function Approximation Error in Actor-Critic Methods},
  author={Scott Fujimoto and Herke van Hoof and David Meger},
  booktitle={International Conference on Machine Learning},
  year={2018},
  url={https://api.semanticscholar.org/CorpusID:3544558}
}

@article{Shi2020TransformerConv,
  title={Masked Label Prediction: Unified Massage Passing Model for Semi-Supervised Classification},
  author={Yunsheng Shi and Zhengjie Huang and Wenjin Wang and Hui Zhong and Shikun Feng and Yu Sun},
  journal={ArXiv},
  year={2020},
  volume={abs/2009.03509},
  url={https://api.semanticscholar.org/CorpusID:221534325}
}

@Article{Mnih2015DQN,
author={Mnih, Volodymyr and Kavukcuoglu, Koray and Silver, David and Rusu, Andrei A. and Veness, Joel and Bellemare, Marc G. and Graves, Alex and Riedmiller, Martin and Fidjeland, Andreas K. and Ostrovski, Georg and Petersen, Stig and Beattie, Charles and Sadik, Amir
and Antonoglou, Ioannis and King, Helen and Kumaran, Dharshan and Wierstra, Daan
and Legg, Shane and Hassabis, Demis},title={Human-level control through deep reinforcement learning},journal={Nature},year={2015},month={Feb},day={01},volume={518},number={7540},pages={529-533},issn={1476-4687},doi={10.1038/nature14236},url={https://doi.org/10.1038/nature14236}
}

@misc{hausknecht2024DDPGmixed,
      title={Deep Reinforcement Learning in Parameterized Action Space}, 
      author={Matthew Hausknecht and Peter Stone},
      year={2024},
      eprint={1511.04143},
      archivePrefix={arXiv},
      primaryClass={cs.AI},
      url={https://arxiv.org/abs/1511.04143}, 
}

@misc{warp2022,
  title        = {Warp: A High-performance Python Framework for GPU Simulation and Graphics},
  author       = {Miles Macklin},
  month        = {March},
  year         = {2022},
  note         = {NVIDIA GPU Technology Conference (GTC)},
  howpublished = {\url{https://github.com/nvidia/warp}}
}

@misc{gurobi,
  author = {{Gurobi Optimization, LLC}},
  title = {{Gurobi Optimizer Reference Manual}},
  year = 2026,
  url = "https://www.gurobi.com"
}

@misc{fey2025pyg,
      title={PyG 2.0: Scalable Learning on Real World Graphs}, 
      author={Matthias Fey and Jinu Sunil and Akihiro Nitta and Rishi Puri and Manan Shah and Blaž Stojanovič and Ramona Bendias and Alexandria Barghi and Vid Kocijan and Zecheng Zhang and Xinwei He and Jan Eric Lenssen and Jure Leskovec},
      year={2025},
      eprint={2507.16991},
      archivePrefix={arXiv},
      primaryClass={cs.LG},
      url={https://arxiv.org/abs/2507.16991}, 
}

@inproceedings{Jingwen2023AutonomousConstruction,
    author = {Wang, J. and Liu, W. and Kao, G. T.-M. and Mitropoulou, I. and Ranaudo, F. and Block, P. and Dillenburger, B.},
    title = {Multi-Robotic Assembly of Discrete Shell Structures},
    booktitle = {Advances in Architectural Geometry 2023},
    year = {2023}
}

@InProceedings{Salamanca2023semiramis,
author="Salamanca, Luis
and Apolinarska, Aleksandra Anna
and P{\'e}rez-Cruz, Fernando
and Kohler, Matthias",
editor="Gengnagel, Christoph
and Baverel, Olivier
and Betti, Giovanni
and Popescu, Mariana
and Thomsen, Mette Ramsgaard
and Wurm, Jan",
title="Augmented Intelligence for Architectural Design with Conditional Autoencoders: Semiramis Case Study",
booktitle="Towards Radical Regeneration",
year="2023",
publisher="Springer International Publishing",
address="Cham",
pages="108--121",
isbn="978-3-031-13249-0"
}

@misc{wang2026learningbuildautonomousrobotic,
      title={Learning to Build: Autonomous Robotic Assembly of Stable Structures Without Predefined Plans}, 
      author={Jingwen Wang and Johannes Kirschner and Paul Rolland and Luis Salamanca and Stefana Parascho},
      year={2026},
      eprint={2602.23934},
      archivePrefix={arXiv},
      primaryClass={cs.RO},
      url={https://arxiv.org/abs/2602.23934}, 
}

\end{document}